\documentclass[nohyperref]{article}

\usepackage{microtype}
\usepackage{graphicx}
\usepackage{subfigure}
\usepackage{booktabs} 

\usepackage{hyperref}

\usepackage[accepted]{icml2022}

\usepackage{amsmath}
\usepackage{amssymb}
\usepackage{mathtools}
\usepackage{amsthm}
 
\usepackage[capitalize,noabbrev]{cleveref}
\usepackage{graphicx}
\usepackage{amsmath}
\usepackage{amssymb}
\usepackage{amsthm}
\usepackage{makecell}
\usepackage{times}
\usepackage{epsfig}
\usepackage{siunitx}
\usepackage{textcomp}
\usepackage{pifont}

\usepackage{dsfont}
\usepackage{algorithm}
\usepackage{algorithmic}
\renewcommand{\algorithmicrequire}{ \textbf{Input:}} 
\renewcommand{\algorithmicensure}{ \textbf{Return}}
\usepackage{array}
\usepackage{booktabs} 
\usepackage{epsfig}
\usepackage{hyperref}
\usepackage{mathrsfs}

\newcommand{\norm}[1]{\left\lVert#1\right\rVert}
\newcommand{\bs}[1]{\boldsymbol{#1}}

\allowdisplaybreaks
\usepackage{listings}
\usepackage{xcolor}
\definecolor{codegreen}{rgb}{0,0.6,0}
\definecolor{codegray}{rgb}{0.5,0.5,0.5}
\definecolor{codepurple}{rgb}{0.58,0,0.82}
\definecolor{backcolour}{rgb}{0.95,0.95,0.92}
\usepackage{multirow}
\usepackage{booktabs}

\DeclareMathOperator*{\argmin}{arg\,min}

\def\bbE{\mathbb{E}}
\def\bbR{\mathbb{R}}
\def\bbP{\mathbb{P}}
\def\bbV{\mathbb{V}}
\def\cR{\mathcal{R}}
\def\cL{\mathcal{R}}

\def\bx{\boldsymbol{x}}
\def\by{\boldsymbol{y}}
\def\bz{\boldsymbol{z}}

\def\bv{\boldsymbol{v}}
\def\bw{\boldsymbol{w}}

\def\cC{\mathcal{C}}
\def\cD{\mathcal{D}}

\def\cL{\mathcal{L}}

\def\cR{\mathcal{R}}

\def\cW{\mathcal{W}}

\def\cX{\mathcal{X}}
\def\cY{\mathcal{Y}}
\def\cZ{\mathcal{Z}}

\usepackage{float}

\crefname{section}{Sec.}{Secs.}
\Crefname{section}{Section}{Sections}
\Crefname{table}{Table}{Tables}
\crefname{table}{Tab.}{Tabs.}
\theoremstyle{plain}
\newtheorem{theorem}{Theorem}[section]

\theoremstyle{definition}
\newtheorem{definition}[theorem]{Definition}
\newtheorem{assumption}[theorem]{Assumption}
\theoremstyle{remark}
\def\bbE{\mathbb{E}}

\def\bbR{\mathbb{R}}
\def\bbP{\mathbb{P}}
\def\bbV{\mathbb{V}}
\def\bx{\boldsymbol{x}}
\def\by{\boldsymbol{y}}
\def\bz{\boldsymbol{z}}

\def\cC{\mathcal{C}}
\def\cD{\mathcal{D}}

\def\cL{\mathcal{L}}

\def\cR{\mathcal{R}}

\def\cW{\mathcal{W}}

\def\cX{\mathcal{X}}
\def\cY{\mathcal{Y}}
\def\cZ{\mathcal{Z}}
\usepackage[textsize=tiny]{todonotes}

\icmltitlerunning{Model Agnostic Sample Reweighting for Out-of-Distribution Learning}

\begin{document}

\twocolumn[
\icmltitle{Model Agnostic Sample Reweighting for Out-of-Distribution Learning}


\icmlsetsymbol{equal}{*}
\begin{icmlauthorlist}
\icmlauthor{Xiao Zhou}{equal,yyy}
\icmlauthor{Yong Lin}{equal,yyy}
\icmlauthor{Renjie Pi}{equal,yyy}
\icmlauthor{Weizhong Zhang}{yyy}
\icmlauthor{Renzhe Xu}{comp}
\icmlauthor{Peng Cui}{comp}
\icmlauthor{Tong Zhang}{yyy,goog}
\end{icmlauthorlist}

\icmlaffiliation{yyy}{The Hong Kong University of Science and Technology}
\icmlaffiliation{comp}{Tsinghua University}
\icmlaffiliation{goog}{Google Research}
\icmlcorrespondingauthor{Tong Zhang}{tongzhang@tongzhang-ml.org}


\icmlkeywords{Machine Learning, ICML, sample reweighting, out-of-domain learning, sparsity, invariant risk minimization, distributional robust optimization}

\vskip 0.3in
]



\printAffiliationsAndNotice{\icmlEqualContribution} 

\begin{abstract}


Distributionally robust optimization (DRO) and invariant risk minimization (IRM) are two popular methods proposed to improve out-of-distribution (OOD) generalization performance of machine learning models.  
While effective for small models, it has been observed that these methods can be vulnerable to overfitting with large overparameterized models. 
This work proposes a principled method, \textbf{M}odel \textbf{A}gnostic sam\textbf{PL}e r\textbf{E}weighting (\textbf{MAPLE}), to effectively address OOD problem, especially in overparameterized scenarios. Our key idea is to find an effective reweighting of the training samples so that the standard empirical risk minimization training of a large model on the weighted training data leads to superior OOD generalization performance. The overfitting issue is addressed by considering a bilevel formulation to search for the sample reweighting, in which the generalization complexity depends on the search space of sample weights instead of the model size. 
We present theoretical analysis in linear case to prove the insensitivity of MAPLE to model size, and empirically verify its superiority in surpassing state-of-the-art methods by a large margin. Code is available at \url{https://github.com/x-zho14/MAPLE}.
\end{abstract}

\vspace{-20pt}
\section{Introduction}
Despite the unprecedented success of deep learning in recent decades, machine learning methods are vulnerable to even slight distributional shift \cite{goyal2019recurrent,sagawa2020investigationdro, gulrajani2020search}. Actually, the common independent and identical distribution (IID) assumption in machine learning can be easily violated due to data selection biases or unobserved confounders that widely exist in real data \cite{liu2021heterogeneous}.  \citet{Arjovsky2019Invariant} suggests that models trained by empirical risk minimization (ERM) can fail to learn causal factors but instead exploit the easier-to-fit spurious correlations, which are prone to distributional shift from training to testing domains \cite{gulrajani2020search}. A typical example is that deep neural networks (DNN) can rely on the background (spurious features: sand or grassland) to distinguish between caw and camel (core features) \cite{beery2018recognition}. Such model can fail dramatically in recognizing a cow in desert. How to enable the deep models to generalize well under distributional shifts is an important long-standing problem.

In an effort to prevent DNN from exploiting the undesired spurious correlation, a popular research direction targets on regularizing DNN during training, including distributionally robust optimization (DRO) \cite{ben2013robust, duchi2019distributionally, duchi2021statistics, sagawa2020investigationdro} and invariant risk minimization (IRM) \cite{Arjovsky2019Invariant, krueger2021outrex, xie2020risk_variance_penalty}. We refer them as \textbf{regularization-based methods} in this paper. DRO aims to optimize the worst case performance in a set of distributions within a certain distance to the original training distribution while IRM tries to learn an invariant representation that discards the spurious features. DRO and IRM have gained their popularity owed to promising performance on small models and datasets \cite{Arjovsky2019Invariant, duchi2019distributionally} and simplicity to perform training in an end-to-end manner. However, they are reported to be less effective when applied to DNNs in recent studies  \cite{sagawa2019distributionally, cherepanova2021technical, lin2021emprical}. Overparamterized DNN can easily reduce the regularization term of DRO or IRM to zero during training while still relying on the spurious features. 

Another line of research is based on reweighting including importance sampling \cite{kanamori2009least, ben2013robust, fang2020rethinking} and stable learning \cite{ kuang2020stable, shen2020stable, xu2021stable}.  We refer to them as \textbf{reweighting-based methods}. They generally perform a two-stage pipeline: 1) reweight the data distribution by some heuristics; 2) perform ERM training on the reweighted distribution. In the first stage, they assign a weight to each sample: importance sampling upweights the rare group inversely to its group size  and stable learning tries to find a weight that makes each feature orthogonal. 
With the weights found in the first stage, the second stage of weighted ERM training becomes resistant to spurious features. Since the first stage is agnostic to the model size of DNN, it does not suffer from the vulnerability of overfitting caused by model overparameterization as in DRO and IRM.
However, the heuristics in those reweighting based methods require more strict prior knowledge like group annoatations to perform well, which makes them less competitive in practice compared with regularization-based counterparts.

In this paper, to resolve the above limitations, we propose a model agnostic sample reweighting method integrating the benefits of two lines of previous works. In short, we solve the overfitting problem of regularization-based methods by taking the weighted ERM training pattern and transform the search space of model parameters into that of sample weights. On the other hand, we avoid the strict requirements of reweighting-based methods by learning sample weights automatically. To achieve this, we formulate the learning of sample reweighting into a bilevel optimization problem. In the inner loop, we train the DNN on the weighted training samples. In the outer loop, we ultilize the OOD criterion evaluated on validation set as the outer objective to guide the learning of the sample weights. We alternatively perform the inner loop and outer loop and finally obtain a set of weights $\bs{w}$ with the such appealing property: with only learnt sample weights and training samples, we are able to perform weighted ERM training to obtain superior OOD performance, without any regularization term or strict prior knowledge on training samples. We use the term \textit{model agnostic} in MAPLE to stress its ability to avoid overfitting 
regardless of the
model size. In addition, the learned sample weights do not have strong dependence on the model used during the searching phase, e.g. the sample weights learned through ResNet-18 can be successfully applied to weighted ERM training on ResNet-50 on the same task (Table \ref{tab:agnostic}).


The general bilevel  framework is presented below:


\vspace*{-10pt}
\begin{itemize}
\item {\textbf{Outer loop.}} Evaluate the model $\bs{\theta}$ by the OOD criterions to measure the model's reliance to spurious features and optimize $\bs{w}$ to minimize the criterion.
\item{\textbf{Inner loop.}} Perform ERM training on the dataset weighted by $\bs{w}$ to obtain learned model $\bs{\theta}$.
\end{itemize}
\vspace*{-10pt}
An appealing feature of this formulation is that the inner loop can be viewed as a mapping from the sample weight space into the DNN parameter space, and the outer loop performs the optimization on weights. Our bilevel optimization framework is less prone to overfitting because it only searches for the weight candidate: the space of weight is much smaller than that of neural networks. For example, CIFAR-10 only contains 50K training data while ResNet-18 exhibits 11.4 million parameters. We empirically demonstrate the effectiveness of MAPLE on various OOD tasks and show that MAPLE surpasses the state-of-the-art methods by a large margin. Remarkably, we achieve even higher worst-group accuracy in Waterbirds without group labels in training samples compared with GroupDRO 
previously recognized as the Oracle upperbound (Table \ref{tab:dro}).



%

Our contributions are summarized as follows:
\vspace{-10pt}
\begin{itemize}
    \item We propose a model agnostic sample reweighting method based on bilevel optimization for OOD learning, which enjoys the following benefits: 
    \begin{itemize}
        \item MAPLE learns sample weights automatically through bilevel optimization avoiding the pathology of conventional reweighting-based methods' reliance on strong prior knowledge on data.
        \item MAPLE transforms the optimization problem from DNN's parameter space to sample weight space, which in turn solves the overfitting problem suffered by regularization-based methods.
    \end{itemize}
    \item We provide theoretical analysis in linear case to prove the existence of ideal sample weight under suitable conditions and insensitivity of the generalization performance to the model capacity of DNN, which is consistent with our empirical results.
    \item We empirically demonstrate the superior performance of MAPLE to state-of-the-art domain generalization methods on various tasks and models.
\end{itemize}

\vspace{-15pt}
\section{Related Work}
\noindent \textbf{Invariant Risk Minimization}. IRM aims to learn a feature representation which elicits a classifier that is simultaneously optimal in various environments \cite{peters2016causal, Arjovsky2019Invariant}.  Several works try to improve IRM by proposing different variants: \cite{krueger2021rex, xie2020risk_variance_penalty} suggest to penalize the variance of the risks among different environments and \cite{chang2020invariant, xu2021learning} try to estimate the invariance violation by training neural networks. \cite{Arjovsky2019Invariant, rosenfeld2020risks,  chen2021iterative} provide theoretical guarantees for IRM on linear models with sufficient training environments. However, IRM is found to be less effective when applied to overparameterized neural networks \cite{gulrajani2020search, yong2021empirical}. \cite{lin2022bayesian} shows that this can be largely attributed to the overfitting problem.  

\noindent \textbf{Distributionally Robust Optimization}. DRO optimizes the worst-case loss in an uncertainty set \cite{ben2013robust, sagawa2019distributionally, duchi2019distributionally,oren2019distributionally, duchi2021statistics, zhang2022towards}. When the uncertainty set is properly chosen, \citet{duchi2019variance, duchi2021learning} shows that DRO can improve the robustness of the learned model by imposing regularization. Unfortunately, similar to IRM, DRO is also shown to be less effective on overparameterized neural networks \cite{sagawa2019distributionally}, which may be largely attributed to the deep model's ability to overfit all the training data.  In an effort to enhance DRO in this case,  \citet{sagawa2019distributionally} suggests to impose large  $\ell_2$ regularization or early stopping on the DNN to alleviate the catastrophic overfitting. \citet{liu2021just} proposes a two-stage method that firstly performs ERM with early stopping and then conduct weighted ERM training by upweighting misclassified samples from the model obtained in the first stage.

\noindent \textbf{Reweighting}.
Sample reweighting is a classic method to deal with distribution shifts. Traditional sample reweighting methods, e.g., importance sampling, assume the prior knowledge of testing distributions are known and they can estimate the density ratio between training and testing distributions directly~\citep{shimodaira2000improving,huang2006correcting,sugiyama2007covariate,sugiyama2008direct,kanamori2009least,fang2020rethinking}. As a result, ERM training on the reweighted distribution is unbiased in the testing distribution~\citep{fang2020rethinking}. Recent works consider a much more challenging setting where the testing distribution is unknown \citep{shen2021towards}. In this direction, stable learning proposes to learn sample weights that make features statistically independent in the reweighted distribution \citep{kuang2020stable,shen2020stable,zhang2021deep, wang2022training, xu2020algorithmic}. \citet{xu2021stable} further theoretically analyze the effectiveness of such algorithms by explaining them as processes of feature selection. However, stable learning is still limited in the sense that  the features need to be provided generally. A recent work aiming at addressing learning with label noise also relies on optimizing sample reweighting using a bilevel framework \cite{ren2018learning}, where a validation set with the same distribution as the test set is needed to ensure good performance. However, in OOD tasks, the training and validation sets are from the same distribution, which is different from the test distribution, rendering these methods inapplicable.

\noindent \textbf{Causality}. The topics covered in this work is closely related to causality.  \citet{peters2016causal} proposes Invariant Causal Prediction (ICP) to utilize the invariance property to identify the direct cause of the target. IRM then extends this idea to DNN by incorporating feature learning \cite{Arjovsky2019Invariant}. Both ICP and IRM need train data to be split into distinct environments, whereas, environments partition is frequently not available in real application. It is of great interest to learn invariance without explicit environment indexes.  \cite{lin2022zin} proposes a framework called ZIN that can provably learn both invariance and environment partition based on the carefully chosen auxiliary information.  DRO is also intrinsically related to causality by noting that causal model optimizes the worst case loss w.r.t. infinite intervention on the causal graph. \cite{rothenhausler2021anchor} explicitly build the connection between distributional robustness with causality. We believe our method is also a potential technique to make causal models compatible with large neural networks. 



\vspace{-2pt}
\section{Preliminaries} \label{sect:basic}
\textbf{Notations}. Given a dataset $\cD := \{(\mathbf{x}_i, \mathbf{y}_i)\}_{i=1}^{n}$ with samples $(\mathbf{x}_i, \mathbf{y}_i)$ drawn from $\mathcal{X} \times \mathcal{Y}$, we  denote weighted empirical loss as $\mathcal{L}(\cD, \bs{\theta}; \bs{w}) := \frac{1}{n}\sum_{i=1}^{n} w_{i}\ell(f(\mathbf{x}_i; \boldsymbol{\theta}), \mathbf{y}_i)$, where $f(\cdot; \bs{\theta})$ is a network  parameterized by $\bs{\theta}$, $\ell(\cdot, \cdot)$ is the loss function, e.g., cross entropy and least square loss, and $w_i\in \mathbb{R}^+$ is the non-negative weight. We denote $\mathcal{L}(\cD, \bs{\theta})$  to be the unweighted loss $\mathcal{L}(\cD, \bs{\theta}; \mathbf{1})$ for abbreviation. Let $\bz_c \in  \mathcal{Z}_c$ and  $\bz_s \in  \mathcal{Z}_s$ be the \textit{core} and \textit{spurious} features.  The core feature is safe to rely on and the reliance on the spurious feature is unstable and unwanted.  We assume the observed feature space is generated by an unknown/known mapping from the core and spurious feature spaces, i.e., $ \mathcal{K}(\cdot, \cdot): \mathcal{Z}_c \times \mathcal{Z}_s \rightarrow \mathcal{X}$.  

IRM and DRO aim to learn a {\it good} predictor $f: \mathcal{X}\rightarrow \mathcal{Y}$, in a sense that $f$ does not rely on the spurious feature $\mathcal{Z}_s$. They formulate it into a minimization problem of different objective functions (referred as {\it OOD Risk}) based on different settings in practice. The details are presented below.
\subsection{IRM}\label{sect:irm_intro}
IRM assumes that we have multiple  environments $\mathcal{E}:=\{e_1, e_2, \ldots, e_E\}$ in the sample space $\cX \times \cY$ with different joint distributions, and the correlation between the spurious features and labels is unstable among different environments. IRM formulates the predictor $f(\cdot; \bs \theta)$ as a composite function of representaion $\phi(\cdot; \Phi)$ and classifier $h(\cdot; \bv)$, i.e., $f(\cdot; \bs \theta) = h(\phi(\cdot; \Phi); \bv) $, where $\bs \theta = \{\bv, \Phi\}$ are the trainable parameters. Its idea is that if a predictor $f(\cdot;\bs{\theta})$ works well on all the environments, then it can be expected that the correlation between the  spurious features  and the labels are not fitted as it is unstable. Therefore, it formulates the task as to minimize a certain OOD risk  to find such good predictor. Two popular 
risks are  
\begin{align}
    \cR^{\textup{IRMv1}}(\cD, \bs{\theta}):=& \sum_{e} \mathcal{L}(\cD^e, \bs{\theta})  + \lambda \| \nabla_v\mathcal{L}(\cD^e, \bs{\theta})\|^2_2\label{irmv1}\\
    \cR^{\textup{REx}} (\cD, \bs{\theta}):=& \sum_{e} \mathcal{L}(\cD^e, \bs{\theta})  + \lambda \bbV_e [ \mathcal{L}(\cD^e, \bs{\theta})]\label{rex}, 
\end{align}
where $\cD = \cup_{e}\cD^e$  with $\cD^e$ being  the  data drawn from environment $e$ and $ \bbV_e [ \mathcal{L}(\cD^e, \bs{\theta})]$ is the variance of the loss across different environments. 

\subsection{DRO}
\label{sect:dro_intro}
DRO aims to optimize the worst case performance in a set of distributions within a certain distance to the original training distribution.

When a set of distributions with different group annotations $g$, i.e., $\cD = \bigcup_g \cD^g$,   is available,  
a popular method named GroupDRO \cite{sagawa2019distributionally} learns a robust predictor by minimizing the following risk, which is actually  the worst-group loss over $\{\cD^g\}_g$, i.e.,:
\vspace{-2pt}
\begin{align}
    \label{Group-DRO}
    \cR_{\textup{Group-DRO}}(\cD, \theta) :=  \max_{g} \cL(\cD^g, \theta).  
\end{align}
\vspace{-2pt}
When such set of distributions is not available, a typical method, conditional value at risk (CVaR) \cite{rockafellar2000optimization},  constructs distributions near the original training distributions by reweighting on the training samples and minimizes a risk defined as the supreme loss over these distributions, i.e., 
\vspace{-2pt}
\begin{align}
    \label{CVaR-DRO}
    \cR_{\textup{CVaR-DRO}}(\cD, \theta) :=  \sup_{\bs w \in \cC(\alpha)} \cL(\cD, \theta; \bs w),  
\end{align}
\vspace{-2pt}
where $\cC(\alpha) = \{\bs{w}: \bs{w} \succeq 0, \|\bs{w}\|_\infty \leq \frac{1} {\alpha n}, \|\bs{w}\|_1=1\}$.




\section{Model Agnostic Sample Reweighting}
In this section, we will first present the bilevel formulation of our proposed MAPLE and provide some theoretical analysis about its generalization ability. Then we will introduce sparsity into MAPLE to enhance its generalization ability.

\subsection{Bilevel Formulation of MAPLE}\label{sect:bi-level}
\begin{figure}
    \centering
    \subfigure[unweighted]
    {\includegraphics[scale=0.26]{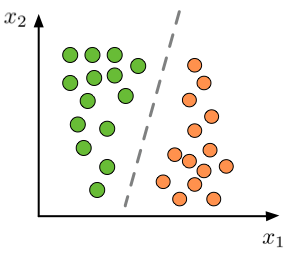}}
    \subfigure[weighted]
    {\includegraphics[scale=0.26]{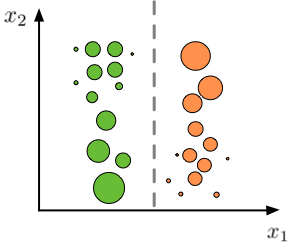}}
    \subfigure[weighted+sparse]
    {\includegraphics[scale=0.26]{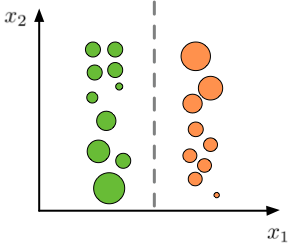}}
    \caption{An illustrative example of removing the reliance on spurious feature via sample reweighting and sparsity constraint on sample size. Circles with larger radius means more weight paid to this training sample. Different colors indicate different labels, i.e., \{0,1\}. Here, $x_1=z_c$ and $x_2= z_s$ are the core and spurious features, respectively.}
    \label{fig:illustration_weight}
\vspace{-4pt}
\end{figure}

We illustrate our key idea using the example in Figure \ref{fig:illustration_weight}, which is to remove the reliance of the learned predictor $f$ on the spurious features by sample reweighting. To be precise, in this example, we assume $x_1$ and $x_2$ are the core and spurious features, and we aim to learn a classifier on these training data. If without reweighting, it is clear that with conventional loss functions, the optimal classifier is the dashed slant line in Figure \ref{fig:illustration_weight}.(a), which depends on $x_2$. If we assign larger weights to the samples in the left-bottem and right-up areas, then the optimal classifier would rotate to be vertical shown in Figure \ref{fig:illustration_weight}.(b). We can see the vertical classifier does not depend on the spurious feature $x_2$, as for fixed $x_1$ and any value of $x_2$, the output of the classifier never changes. Therefore, it shows that we can remove the reliance on the spurious features by sample reweighting. Thus, the problem comes to how to automatically learn appropriate weights for training samples.

Consider a training dataset $\cD_{tr}:=\{(\mathbf{x}^{tr}_i, \mathbf{y}^{tr}_i)\}_{i=1}^{n_{tr}}$ and a validation dataset $\cD_{v}:=\{(\mathbf{x}^{v}_i, \mathbf{y}^{v}_i)\}_{i=1}^{n_{v}}$ randomly partitioned from dataset $\cD$. We formulate the task of learning sample weights to remove the reliance on the spurious features as the following  bilevel optimization problem:
\begin{gather}
\min_{\bs{w} \in \mathcal{C}}  \cR(\cD_{v}, \boldsymbol{\theta}^{*}(\bs{w})) , \label{ori_formulation}\\
s.t. ~\boldsymbol{\theta}^{*}(\bs{w}) \in \argmin_{\boldsymbol{\theta}} \mathcal{L}(\cD_{tr}, \bs{\theta}; \bs w) , \label{ori_formulation_outer_loop} 
\end{gather}
where $\bs{w}$ is a sample weight vector with length $n_{tr}$ indicating the importance of training samples, $\mathcal{C} = \{\bs{w}: \bs{w} \succeq 0\}$. Any OOD Risk $\cR(\cD, \boldsymbol{\theta})$ described in the Section \ref{sect:basic} can be used as the outer objective here.  In the inner loop, we minimize the weighted ERM loss on training samples, in order to obtain a model $\boldsymbol{\theta}^{*}(\bs{w})$, and in the outer loop,  we evaluate the learned model's reliance on spurious features through OOD Risk and optimize sample weights. By alternatively performing inner loop and outer loop, the sample weights gradually evolve to the state of being able to produce satisfactory OOD performance with simply ERM training.

 
Moreover, our formulation has the following advantages:
\begin{itemize}
    \item In our framework, we essentially define an implicit mapping from the sample weight space to the model parameter space, which enables us to learn in the sample weight space. As the sample weight space is always significantly smaller than model parameter spaces, we can avoid the pathology of overfitting caused by overparameterization.  
    \item Compared with existing regularization-based methods, MAPLE adopts validation dataset in the outer loop to alleviate the problem of overfitting to training dataset.  
\end{itemize}
These advantages are consistent with our theoretical analysis (Section \ref{ana}) and  empirical observations (Section \ref{sec:exp}).

\textbf{[Optimization by Truncated Back-propagation and Projected Gradient Descent].} The above bilevel optimization can be solved by performing  projected gradient descent to $\bs{w}$. The gradient of $\bs{w}$ can be calculated by:
{\small
\begin{align}
&~~~~~~\nabla_{\bs{w}} \cR \nonumber\\ 
&= \left.\nabla_{\bs{\theta}}\cR\right|_{\theta^{*}}\nabla_{\bs{w}}\boldsymbol{\theta}^{*}\label{1}\\ 
                    &\approx \left.\nabla_{\bs{\theta}}\cR\right|_{\theta_{T}}\nabla_{\bs{w}}\boldsymbol{\theta}_{T} \label{2}\\ 
                    &= \left.\nabla_{\bs{\theta}} \cR \right|_{\theta_{T}}\left.\sum_{ j \leq L}\left[\prod_{k<j} I-\left.\frac{\partial^{2} \mathcal{L}}{\partial \bs{\theta} \partial \bs{\theta}^{\intercal}}\right|_{\bs{\theta}_{T-k-1}}\right] \frac{\partial^{2} \mathcal{L}}{\partial \bs{\theta} \partial \boldsymbol{w}^{\intercal}}\right|_{\bs{\theta}_{T-j-1}} \nonumber \\ 
                    &\approx \left.\nabla_{\bs{\theta}} \cR\right|_{\theta_{T}}\left.\frac{\partial^{2} \mathcal{L}}{\partial \bs{\theta} \partial \boldsymbol{w}^{\intercal}}\right|_{\bs{\theta}_{T-1}} \label{3},
\end{align}
}
where Eqn. (\ref{1}) follows chain rule, Eqn. (\ref{2}) approximates $\theta^{*}$ by $\theta_{T}$ obtained from $T$ steps of inner loop gradient descent and Eqn. (\ref{3}) performs 1-step truncated backpropagation \cite{shaban2019truncated}. Then MAPLE updates $\bs{w}$ by projected gradient descent:
\begin{align}
\bs{w} \leftarrow \operatorname{proj}_{\mathcal{C}}\big(\bs{w} - \eta\left.\nabla_{\bs{\theta}} \cR\right|_{\theta_{T}}\left.\frac{\partial^{2} \mathcal{L}}{\partial \bs{\theta} \partial \boldsymbol{w}^{\intercal}}\right|_{\bs{\theta}_{T-1}}\big), 
\end{align}
where $\eta$ is the learning rate.


\subsection{Theoretical Analysis on Linear Case} \label{ana}
In this section, we analyze the performance of our method in the linear case where we consider Problem \eqref{ori_formulation} with linear predictor $f(\bx; \bs \theta) = \bx^\intercal \bs \theta, \bs \theta \in \bbR^d$ and least square loss $\ell(f(\bx), \by) = \|f(\bx)-\by\|^2_2 $. We further consider $\bx \in \bbR^d$ to be the generated from core features $\bz_c \in \bbR^{d_c}$ and spurious features $\bz_s \in \bbR^{d_s}$ by a transformation matrix $\bs S \in \bbR^{d \times (d_c+d_s)}$, i.e., $\bx = \bs S [\bz_c; \bz_s]$. We assume $d_c+d_s=d$ for simplicity and assume the feature transformation $\bs S$ is invertible by some matrix $\bs T \in \bbR^{ (d_c+d_s) \times d}$ such that $\bs T \bs S ([\bz_c; \bz_s]) = [\bz_c; \bz_s]$. Our goal is to learn a  function  $f$ that predicts $\by$ based on $\bx$ without reliance on $\bz_s$. Let $\bbP(\bx, \by)$ denote the distribution on the training and validation sets as defined in Section \ref{sect:bi-level}. We further use $\bbE$ to denote the expectation w.r.t. $\bbP(\bx, \by)$.

In Section \ref{sect:population_property}, we consider the population level property, i.e., when infinite samples are available. In Section \ref{sect:finite_property}, we consider the case with finite samples.

\subsubsection{Population Level Properties}
\label{sect:population_property}
At first, we need to extend  the  weight and loss of problem \eqref{ori_formulation_population} into the population level as follows.

\begin{definition} 
\label{defi:wegihting_function}
We define the set of  weight functions as  
\begin{align*}
    \cW = \{ w: \cX \times \cY \xrightarrow[]{} \bbR^{+}| \bbE[w(\bx, \by)] = 1\}.
\end{align*}
Given any $w \in \cW$, the populated unweighted  and weighted loss can be defined as 
\begin{align}
    \cL(\bs \theta)  &= \int (y - \bx^\intercal \bs \theta)^2 \bbP(\bx, y) d \bx dy  \\
     \cL(\bs \theta; w)  &= \int (y - \bx^\intercal \bs \theta)^2 \bbP_w(\bx, y) d \bx dy \label{eqn:weighted_loss},
 \end{align}
where $\bbP_w(\bx, \by) = w(\bx, \by)\bbP(\bx, \by)$ is the weighted distribution. 
\end{definition}
The populated version of problem \eqref{ori_formulation} takes the form of 
\begin{gather}
\min_{\bs{w} \in \mathcal{C}}  \cR(\boldsymbol{\theta}^{*}({w})) , 
\label{ori_formulation_population}\\
s.t. ~\boldsymbol{\theta}^{*}({w}) \in \argmin_{\boldsymbol{\theta}} \mathcal{L}(\bs{\theta}; w) ,  \label{eqn:population_theta_sol}
\vspace{-4pt}
\end{gather}
here $\cR(\bs \theta)$ is the populated OOD risk obtained by replacing the empirical loss with the populated one in Eqn \eqref{irmv1}-\eqref{CVaR-DRO}. We assume  the solution in the inner loop is unique.  We define the optimal linear classifier as the one that minimizes the expected loss without using any spurious features:

\begin{definition}
\label{defi:optimal_debiased_predictor}We define  the optimal debiased predictor as 
$$ \bs {\bar \theta} :=  \bs T^\intercal[\bs {\bar \theta}_c; \mathbf{0}],$$ where
$\bs{\bar \theta}_c  := \argmin_{\bs{\theta}_c}  \bbE[\|  y - \bz^\intercal_c \bs{\theta}_c\|^2].$
\end{definition}

We now make further assumptions as follows:
\begin{assumption}[Strictly positive density]
\label{ass:strict_pos}
 $\forall \mathbf{y} \in \cY, \mathbf{z}_c \in \cZ_c, \mathbf{z}_s \in \cZ_s$, $P(\bz_c=\mathbf{z}_c, \bz_s=\mathbf{z}_s, \by=\mathbf{y}) > 0$.
\end{assumption}

\begin{assumption}
\label{ass:identifibility}
The optimal debiased predictor $\bs{\bar \theta}$ is identifiable by the populated OOD Risk $\cR$, i.e.,
\begin{align*}
    \cR(\bs{\bar \theta}) < \cR(\bs{\theta}), \forall \bs{\theta} \in \bbR^d, \bs{\theta} \neq \bs{\bar \theta}.
\vspace{-5pt}
\end{align*}
\end{assumption}
Assumption \ref{ass:strict_pos} is common in existing works because there always exists uncertainty in the data  \cite{pearl1988probabilistic,strobl2016markov,xu2021stable}. Assumption \ref{ass:identifibility}  is a natural condition, making it possible to provably identify $\bs{\bar \theta}$ by using $\cR$. For example, it has  been demonstrated that the metrics of IRM  can satisfy this condition with sufficient number of  environments \cite{Arjovsky2019Invariant, rosenfeld2020risks}.  

\begin{theorem}
[Identifiability on population level] 
\label{thm:Identifiability}
When Assumption \ref{ass:strict_pos} holds, there exists a weight function $w\in \cW$, such that the optimum solution of Eq. \eqref{eqn:population_theta_sol} satisfies that
$$\bs{\theta^*}(w) = \bs {\bar \theta}.$$ Further, when Assumption \ref{ass:identifibility} holds, the populated MAPLE, i.e., Eqn.\eqref{ori_formulation_population}-\eqref{eqn:population_theta_sol}, can uniquely identify $\bs { \bar \theta}$.
\end{theorem}
The theorem above shows MAPLE can provably find the sample weight to removes reliance of model on the spurious features. This verify the main idea illustrated in Figure \ref{fig:illustration_weight}.

\subsubsection{Finite Sample}
\label{sect:finite_property}
Now we turn to analyze the finite sample case. By extending the weight vector $\bs w$ into the functional form $w(\bx, y)$ in Definition \ref{defi:wegihting_function}, we rewrite problem  \eqref{ori_formulation} into:
\begin{gather}
\min_{\bs{w} \in \mathcal{C}}  \cR(\cD_{v},\bs{\hat \theta}^{*}({w})) , \label{eqn:theretical_finite_maple}\\
s.t. ~ \bs{\hat \theta}^{*}({ w}) = \argmin_{\boldsymbol{\theta}} \mathcal{L}(\cD_{tr}, \bs{\theta};  w) ,  \nonumber
\end{gather}
Then given a weight function  $w$, $\hat{\bs\theta}^*(w)$  is a deterministic mapping from $\cW$ to the parameter space. Suppose we can find a $\hat w$ that is a $\epsilon-$approximate solution of minimizing $\cR(\cD_{v},\bs{\hat \theta}^{*}({w}))$, i.e.,
\begin{align}
\label{eqn:epsilon}
\cR(\cD_{v},\bs{\hat \theta}^{*}({\hat w})) \leq \inf_{w \in \cW} \cR(\cD_{v},\bs{\hat \theta}^{*}({w})) + \epsilon.
\end{align}
Observing that $\bs{\hat \theta}^{*}(\cdot)$ only depends on $\cD_{tr}$, we can obtain the following generalization bound with standard uniform convergence analysis on $\cD_{v}$:

\begin{theorem}[Finite Samples]
\label{thm:finite_bound}
Suppose $|\cD_{v}| =n$. Let $\cD_v^{-1}$ denote the dataset generated by replacing one sample in $\cD_v$ with another arbitrary sample. Assume there exists a constant $M>0$ such that $\forall \bs \theta, |\cR(\cD_{v}, \bs \theta) - \cR(\cD_{v}^{-1}, \bs \theta)| \leq M/n,  $ where $\cR(\cD, \bs \theta)$ denotes the OOD risk on the dataset $\cD$. Further assume $\cW$ contains $|\cW|$ discrete choices.   With probability at least $1-\delta$, MAPLE outputs a solution $\hat w$ satisfies
\begin{align}
    \label{eqn:finite_bound}
    \cR(\bs{\hat \theta}^{*}( \hat {w})) &\leq \inf_{w \in \cW } \cR(\bs{\hat \theta}^{*}({w})) + \epsilon + M\sqrt{\frac{2\ln(2|\cW|/\delta)}{n}},
\end{align}
where $\cR(\bs \theta)$ is the populated OOD risk achieved by $\bs \theta$, $\bs{\hat \theta}$ and $\epsilon$ are defined in Eqn. \eqref{eqn:theretical_finite_maple} and Eqn. \eqref{eqn:epsilon}, respectively. 
\end{theorem}
Theorem \ref{thm:finite_bound} shows that the generalization performance depends on the complexity of $\cW$ and the size of validation dataset. As our weight space $\cW$ is usually significantly smaller than the parameter space, MAPLE could have  better generalization performance compared with training OOD risk directly on DNN. Further, the RHS of Eqn. \eqref{eqn:finite_bound} does not involve  the complexity of neural networks, indicating MAPLE is insensitive to the model size. Extensive experimental results in Section \ref{sec:exp} verify this result, showing MAPLE can achieve significant better performance than existing methods, especially on large models. We'd like to point out that Theorem \ref{thm:finite_bound} still holds when $ \bs{\hat \theta}^*(\cdot)$ is a general non-linear function because the theorem is a direct application of the standard uniform convergence analysis which doesn't require $\bs{\hat \theta}^*(\cdot)$ to be linear. 

\begin{algorithm*}[htb!]
\caption{Model Agnostic Sample Reweighting (MAPLE)}
\label{alg:SST}
\begin{algorithmic}[1]
\REQUIRE a network $\bs{\theta}$, remaining training sample size $K$, training set $\mathcal{D}_{tr}$ and validation set $\mathcal{D}_{v}$.
\STATE Initialize sample weights $\bs{w}=\mathbf{1}$ and probabilities $\bs{s}=\frac{K}{|\mathcal{D}_{tr}|}\mathbf{1}$.
\FOR{ training iteration $i = 1, 2 \ldots I$}
\STATE Sample mask $\bs{m}$ according to the probability distribution $p(\boldsymbol{m}|\boldsymbol{s})=\Pi_{i=1}^{n} (s_i)^{m_i}(1-s_i)^{(1-m_i)}$. 
\STATE Train the inner loop to converge: $\boldsymbol{\theta}^{*}(\bs{w}, \bs{m}) \leftarrow \argmin_{\boldsymbol{\theta}} \mathcal{L}(\mathcal{D}_{tr}, \bs{\theta}; \bs{w}, \bs{m})$ started from randomly initialized $\bs{\theta}$.
\STATE Estimate $\nabla_{\bs{s}}\Phi(\bs{w}, \bs{s})$ and $\nabla_{\bs{w}}\Phi(\bs{w}, \bs{s})$ by Straight-through Gumbel-softmax and 1-step truncated backpropagation. 
\STATE Perform projected gradient descent: $(\bs{w},\bs{s}) \leftarrow  \operatorname{proj}_{\cC'} ({\bs{w}-\eta\nabla_{\bs{w}}\Phi(\bs{w}, \bs{s}),\bs{s}-\eta\nabla_{\bs{s}}\Phi(\bs{w}, \bs{s})})$

\ENDFOR
\OUTPUT The weighted set $\{(\mathbf{x}_i, \mathbf{y}_i, w_i): m_i \neq 0, {(\mathbf{x}_i, \mathbf{y}_i)} \in \cD_{tr}\}$ with $\bs{m}$ sampled from $p(\boldsymbol{m}|\boldsymbol{s})$ 
\end{algorithmic}
\end{algorithm*}

\subsection{Enhance MAPLE by sparsity}


As shown in Figure \ref{fig:illustration_weight}.(c), we further impose a sparsity constraint on the training sample size, i.e., $\mathcal{C}~ \text{becomes}~ \{\bs{w}: \bs{w} \succeq 0, \norm{\bs{w}}_0 \leq K\}$ in order to save the computational cost in the inner loop. We will verify the benefit in our experiment. Intuitively, sparsity can be seen as forcing several sample weights to be zero. In this way, noisy data samples are removed. Inspired by previous works on $L_0$ regularization optimization \cite{louizos2018learning, zhou2021efficient, zhou2021effective, zou2019sufficient}, we relax the original formulation to be continuous:
\begin{gather}
\min_{(\bs{w}, \bs{s})\in \mathcal{C'}} \Phi(\bs{w}, \bs{s}) = \displaystyle \mathbb{E}_{ p(\boldsymbol{m}|\boldsymbol{s})} ~ \cR(\cD_{v}, \boldsymbol{\theta}^{*}(\bs{w}, \bs{m})), \label{cont_formulation}\\
s.t. ~\boldsymbol{\theta}^{*}(\bs{w}, \bs{m}) \in \argmin_{\bs{\theta}} \mathcal{L}(\cD_{tr}, \bs{\theta}; \bs{w}\circ\bs{m}) \nonumber
\end{gather}
where $\mathcal{C'} = \{(\bs{w}, \bs{s}): \bs{w} \succeq 0, 0 \preceq \bs{s} \preceq 1, \norm{\bs{s}}_1 \leq K\}$ is the feasible set, $m_i$ is viewed as a Bernoulli random variable with probability $s_i$ to be $1$ and $1-s_i$ to be $0$. Assuming the variables $m_i$ are independent, we can get $p(\boldsymbol{m}|\boldsymbol{s}) = \Pi_{i=1}^{n} (s_i)^{m_i}(1-s_i)^{(1-m_i)}$. The discrete constraint $\|\boldsymbol{w}\|_0\leq K $ in problem (\ref{ori_formulation}) can be  relaxed into $\norm{\bs{s}}_1 \leq K$. 

We calculate the gradient to $\bs{w}$ and $\bs{s}$ by Straight-through Gumbel-softmax \cite{paulus2021raoblackwellizing}:
\begin{align}
    &\nabla_{\bs{w}, \boldsymbol{s}}\Phi \approx \nabla_{\bs{w}, \boldsymbol{s}} \cR(\boldsymbol{\theta}^{*}(\boldsymbol{w}, \mathds{1}(\log(\frac{\boldsymbol{s}}{\boldsymbol{1}-\boldsymbol{s}})+\boldsymbol{g_1}-\boldsymbol{g_0} \geq 0))),\nonumber
\end{align}
where $\boldsymbol{g_0}$ and $\boldsymbol{g_1}$  are two random variables with each element IID sampled from $\operatorname{Gumbel}(0,1)$ and the following calculations are similar to those of Eqn. \ref{3}. Then MAPLE updates $\bs{w}$ and $\bs{s}$ by projected gradient descent:
\begin{align}
(\bs{w}, \bs{s}) \leftarrow \operatorname{proj}_{\mathcal{C'}}(\bs{w} - \eta\nabla_{\bs{w}}\Phi, \bs{s} - \eta\nabla_{\bs{s}}\Phi), 
\end{align}
where $\eta$ is the learning rate.

\section{Experiment}\label{sec:exp}
In this section, we conduct a series of experiments to justify the superiority of our MAPLE in IRM and DRO. Detailed dataset descriptions and experimental configurations are placed in appendix due to space limitation. 


\begin{figure*}[t!]
\begin{center}
\centering  
\includegraphics[scale=0.5]{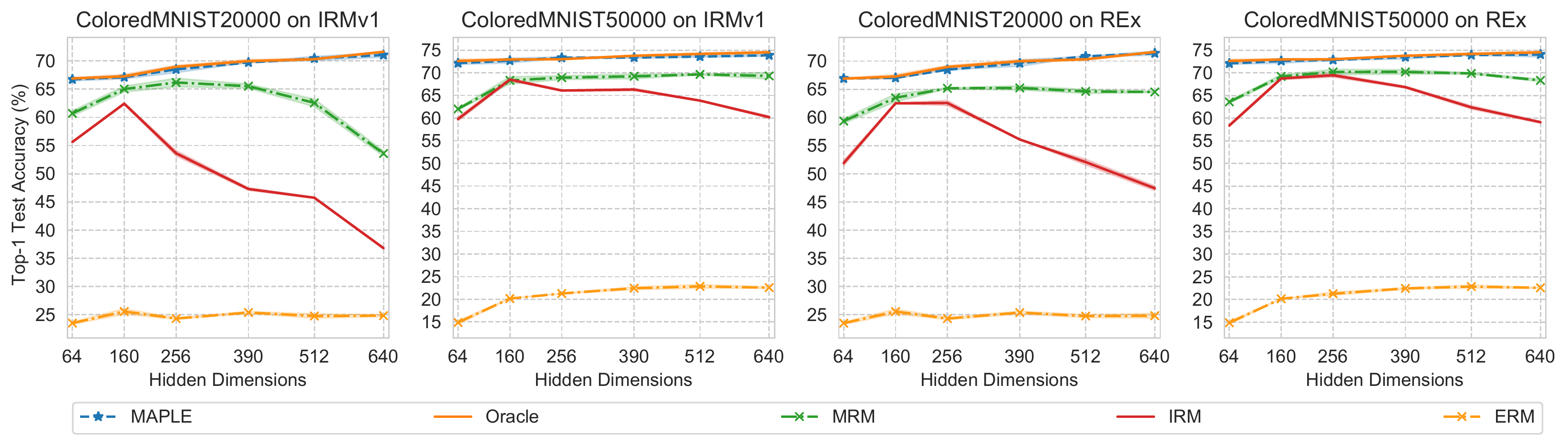}
\end{center}
\caption{Comparing MAPLE with Oracle, IRM, MRM \cite{zhang2021can} and ERM on MLP on ColoredMNIST with varying hidden dimensions and dataset sizes. Oracle is the method done with ERM training with no spurious features and serves as an upper bound. The left (right) two figures demonstrate the comparison of MAPLE with IRMv1 (REx) where MAPLE adopts the same IRMv1 (REx) loss as the outer objective. MAPLE achieves comparable generalization performance with Oracle in all settings.}\label{fig:MLPCMNIST}
\end{figure*}

\subsection{Datasets and Baselines}
\textbf{[Datasets]}.  For IRM experiments, ColoredMNIST is the most widely used benchmark in IRM and ColoredObject, CIFARMNIST are adopted to showcase the superior performance of MAPLE on more challenging largescale settings \cite{Arjovsky2019Invariant, krueger2021rex, ahuja2020invariant, zhang2021can}. We adopt two popular vision datasets, Waterbirds and CelebA, to validate the effectiveness of MAPLE on DRO problems \cite{WahCUB_200_2011, sagawa2019distributionally,liu2015deep, sagawa2019distributionally, liu2021just, yong2021empirical}. Waterbirds and CelebA are real-world datasets and we adopt them to demonstrate the generalizability of MAPLE to real-world scenarios. We follow the challenging setting of \citet{liu2021just} where no group annotation is provided in the training dataset.

\textbf{[Baselines]}. To demonstrate the superiority of our MAPLE on IRM, we compare with standard empirical risk minimization (ERM), two popular foundational invariant risk minimization methods IRMv1 \cite{Arjovsky2019Invariant}  and REx \cite{krueger2021rex} and the latest competitive method MRM \cite{zhang2021can} and SparseIRM \cite{sparseirm} which boost IRM via imposing sparsity. SparseIRM imposes sparsity during training while MRM imposes sparsity after training. We also compare with  BayesianIRM \cite{lin2022bayesian} which introduces Bayesian Inference into IRM to estimate a distribution of classifiers. We include ERM trained on datasets without spurious features to serve as an upper bound (Oracle). To showcase the effectiveness of MAPLE on DRO, we compare with standard empirical risk minimization (ERM), three widely-used DRO methods without group annotations on the training samples: CVaR DRO \cite{levy2020largescale} which is described in Eqn. \eqref{CVaR-DRO}, Learn from failure (LfF)  \cite{nam2020learning}, Just Train Twice (JTT) \cite{liu2021just} and two DRO methods demanding group annotations on the training samples: UpWeighting \cite{cui2019classbalanced, cao2019learning}, GroupDRO \cite{sagawa2019distributionally}.

\subsection{MAPLE on IRM}
\label{sect:results_mlp_mnist}

\textbf{[IRM on ColoredMNIST]}. For different vanilla IRM method (IMRv1 or REx) to be compared, we ultilize the same IRM loss as the outer objective in MAPLE.  We vary the number of training sample size and model parameters to demonstrate the general applicability to various scales in practice.  We add a number to the end of the dataset name to indicate the training set size. We split 10\% training data as the validation dataset.


From Figure \ref{fig:MLPCMNIST}, vanilla IRM methods still lags behind the Oracle performance by a large margin. The gap becomes more prominent when the model is more overparameterized. MRM further boosts the generalization performance while its performance is still limited by regularization-based IRM training paradigm. MAPLE transforms the search space of model parameters into that of sample weights and searches for the optimal sample weights on training dataset, and further guides the optimization by evaluating the criterion on the learned model. MAPLE beats these latest competitive baselines by a large margin and even approaches the performance of Oracle.

\begin{table}[htb!]
\caption{Comparison of Top-1 Test Accuracy on ResNet-18 on ColoredObject and CIFARMNIST.}\label{tab:ResCOCO}
\begin{center}
{\footnotesize
\begin{tabular}{p{1.2cm}<{\centering} p{1.2cm}<{\centering} p{1.7cm}<{\centering}p{1.7cm}<{\centering}}
\toprule
\multicolumn{2}{c}{Dataset}& ColoredObject & CIFARMNIST \\ \cmidrule(){1-4}
\multicolumn{2}{c}{Oracle} & $87.9 \pm 0.3$  & $83.7 \pm 1.5$\\  \cmidrule(r){1-2} \cmidrule(l){3-4}
\multicolumn{2}{c}{ERM} & $51.6 \pm 0.5$ & $39.5 \pm 0.4$ \\ \midrule[0.6pt]
\multicolumn{2}{c}{BayesianIRM} & $78.1 \pm 0.6$ & $59.3 \pm 0.8$ \\ \midrule[0.6pt]
&IRM & $72.5 \pm 2.1$ & $51.3 \pm 3.0$ \\  \cmidrule(r){2-2} \cmidrule(l){3-4}
IRMv1b & MRM & $58.4 \pm 0.9$ & $56.7 \pm 2.3$ \\
\cmidrule(r){2-2} \cmidrule(l){3-4}
 & SparseIRM & \textbf{$87.4\pm 0.6$} & $63.9 \pm 0.4$ \\
\cmidrule(r){2-2} \cmidrule(l){3-4}
& MAPLE & \textbf{87.4 $\pm$ 0.5} & \textbf{82.9 $\pm$ 0.4}  \\ \midrule[0.6pt]
&IRM  & $73.8 \pm 1.3$ & $50.1 \pm 2.2$ \\  \cmidrule(r){2-2} \cmidrule(l){3-4}
REx&MRM & $55.7 \pm 2.9$ & $52.6 \pm 1.5$ \\
\cmidrule(r){2-2} \cmidrule(l){3-4}
 & SparseIRM & $80.3 \pm 1.1$ & $62.7 \pm 0.6$\\
 \cmidrule(r){2-2} \cmidrule(l){3-4}
&MAPLE &\textbf{86.9 $\pm$ 1.0}  & \textbf{82.5 $\pm$ 0.7} \\  \bottomrule
\end{tabular}
}
\end{center}
\vskip -0.1in

\end{table}

\begin{table*}[htb!] 
\caption{Comparison of MAPLE and state-of-the-art DRO methods in Waterbirds and CelebA. MAPLE surpasses previous methods without group annotations by a large margin and even achieves comparable or even better performance than GroupDRO and Upweighting, which utilize the group annotation for training samples.  }
\begin{center}
{\footnotesize
\begin{tabular}{m{4.5cm}<{\centering} m{2.8cm}<{\centering} m{1.7cm}<{\centering}m{1.7cm}<{\centering}m{1.2cm}<{\centering} m{1.7cm}<{\centering}m{1.7cm}<{\centering}}
\toprule
Method& Group annotations for training samples?& \multicolumn{2}{c}{Waterbirds} & \multicolumn{2}{c}{CelebA} \\ \cmidrule(){1-6}
& & Average & Worst-group & Average & Worst-group\\\cmidrule(){3-6}
Upweighting \cite{cui2019classbalanced} & Yes & 92.2 & 87.4 & 89.3 & 83.3  \\ 
GroupDRO \cite{sagawa2019distributionally}  & Yes & 93.5 & 91.4 & 92.9 & 88.9 \\   \cmidrule(){1-6}
ERM & No & 97.3 & 72.6 & 95.6 & 47.2 \\ 
CVaR DRO \cite{levy2020largescale}  & No & 96.0 & 75.9 & 82.4 & 64.4\\
LfF \cite{nam2020learning} & No & 91.2 & 78.0 & 86.0 & 70.6 \\
JTT \cite{liu2021just} & No & 93.3 & 86.7 & 88.0 & 81.1\\ 
MAPLE & No & \textbf{92.9} & \textbf{91.7} & \textbf{89.0} & \textbf{88.0} \\  
\bottomrule
\end{tabular}
}
\end{center}
\label{tab:dro}
\end{table*}

\textbf{[IRM on ColoredObject and CIFRAMNIST]}. In this section, we evaluate the performance of MAPLE on ColoredObject and CIFARMNIST with large-sized model ResNet-18 in Table \ref{tab:ResCOCO}. We split 10\% training data as the validation dataset. MAPLE consistently beats the baselines by a large margin and achieves performance approaching Oracle. These results validate the effectivenss of MAPLE on more modern ResNet architecture and diverse tasks. Notably MAPLE surpasses vanilla IRM method by over 30\% percent in the CIFARMNIST dataset. It 
shows that in more challenging scenarios MAPLE can outperform IRM by a larger margin. 


\subsection{DRO on Waterbirds and CelebA}

In this section, we further validate the effectiveness of MAPLE when applied to DRO. 
The worst-group accuracy is taken as the core criterion to evaluate the effectiveness of DRO methods. In this experiment, we adopt the CVaR DRO objective. To be noted, our bilevel formulation doesn't rely on the group annotations on training samples. We set the $\alpha$ to be 20\% and we find it serves as a good threshold without hyperparameter search.

From the Table \ref{tab:dro}, we find that MAPLE beats previous state-of-the-art method JTT without group annotations on training samples by 5\% in Waterbirds and 6.9\% in CelebA. This can be expected as JTT upweights mis-classified training samples by a mannually-searched magnitude, by evaluating a checkpoint obtained from ERM training at a manually-searched epoch. This inevitably leads to suboptimal performance due to its cumbersome criterion of just upweighting the misclassified training samples at a specific epoch rather than considering more globally is imperfect. To be totally contrary, MAPLE ultilizes the CVaR DRO criterion evaluated on validation set to consider the problem more reasonably by gradually optimizing the sample weights through evaluating the model learned from current sample weights step by step. Upweighting simply upweights the rare groups inversely to its portion in the whole dataset and ignores the importance differed from sample to sample in the same group. GroupDRO makes further improvement to Upweighting by regularziation term and is generally considered as a upperbound by previous works \cite{liu2021just}. MAPLE surpasses Upweighting by 4.3\% and GroupDRO by 0.3\% in Waterbirds and surpasses Upweighting by 4.7\% in CelebA demonstrating the effectiveness of MAPLE in more complex DRO setting.




\begin{figure}[t!] 
\begin{center}
\centering  
\includegraphics[scale=0.5]{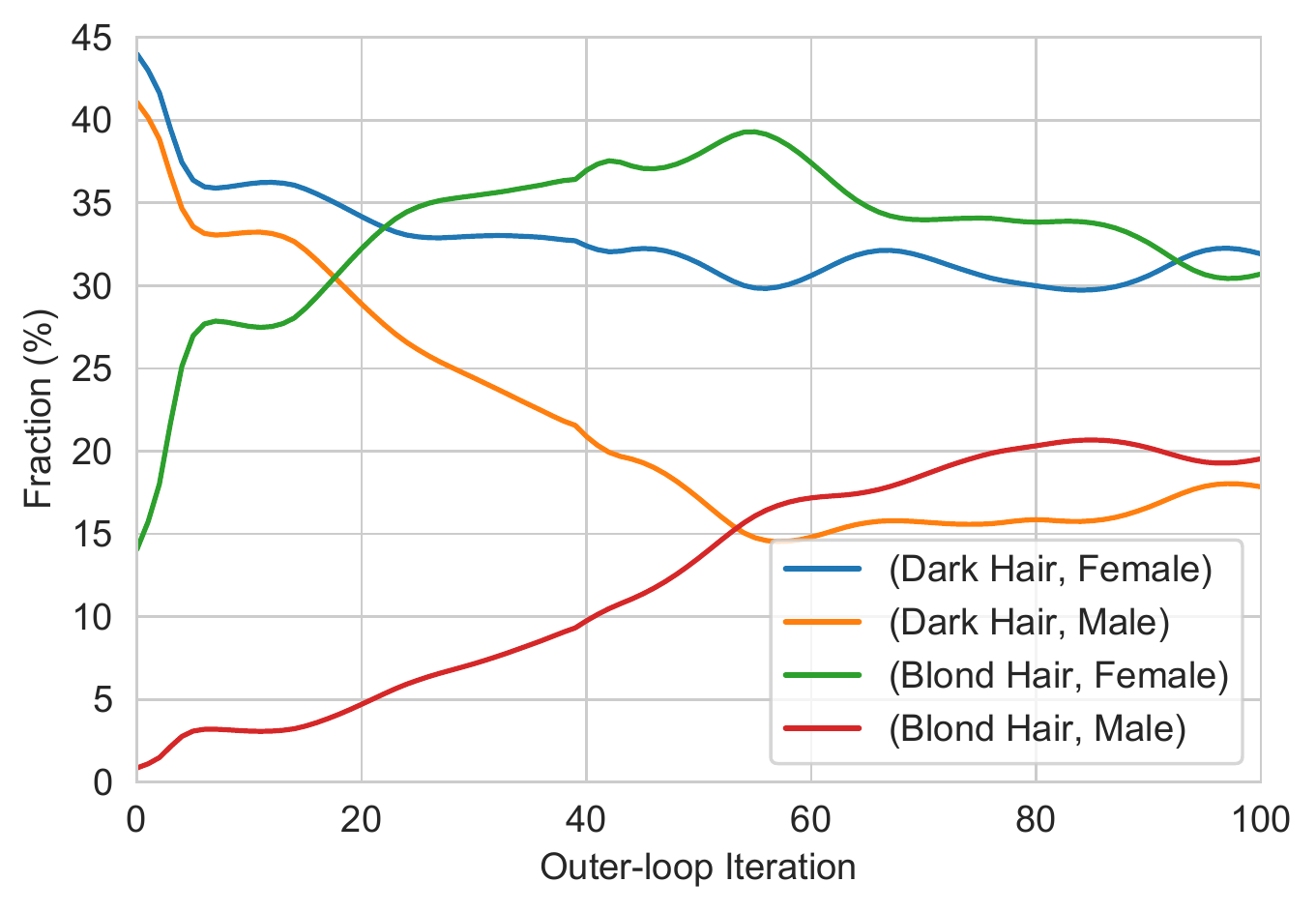}
\end{center}
\caption{Training dynamics of each group weight fraction for ResNet-50 on CelebA. The weight fraction of (Blond Hair, Male) and (Dark Hair, Male) changes to 20\%. The weight fraction of (Dark Hair, Female) and (Blond Hair, Female) changes to 30\%. This indicates that MAPLE can automatically adjust the weight fraction of different groups and the weight fraction of four groups need not be the same.}
\vspace*{-5pt}
\label{fig:ablation}
\end{figure}

\subsection{Further Analysis}\label{sec:ablation-study}

\textbf{[Training dynamics of the weights of different groups]} We plot the training dynamics of sample weight fraction in CelebA experiment in Figure \ref{fig:ablation}. Initially all the weights of different samples are initilized as 1. As there are scarce training samples in group (Blond Hair, Male), its weight fraction is initially only 0.085\%. After 100 iterations of updates, the weight fraction of (Blond Hair, Male) gradually comes up to approximately 20\%. Concurrently, the weight fraction of group (Dark Hair, Male) goes down to approximately 20\% and the weight fraction of (Dark Hair, Female) and (Blond Hair, Female) both come to approximately 30\%. This demonstrates that we need not upweight each group to the same importance level, which indicates one reason why Upweighting fails behind MAPLE.

\textbf{[Weight Distributions of Four Groups]} We further plot the histogram of samples weights in Figure \ref{fig:hist} for four groups in CelebA experiment at the end of training. It indicates that the weights of group (Blond Hair, Female) flattens to around 30. This is consistent with our primal goal to upweight the group with few training samples, and MAPLE sucessfully achieve this without any training group annotations. We also discovers that the sample weight assigned to different training samples need not be the same. This demonstrates another reason why MAPLE beats JTT and Upweighting by a large margin.





\section{Conclusion}
In this work, we present a model agnostic sample reweighing method named MAPLE for out-of-domain learning. We propose a novel bilevel optimization framework to learn sample weights to address the out-of-domain learning problem effectively. We further enhance MAPLE with sparsity to improve training speed. We present theorectical analysis in linear case and demonstrate its superior performance various tasks and models.
\section*{Acknowledgements}
XZ, YL, RP, WZ and TZ acknowledge the funding supported by GRF 16201320. RX and PC acknowledge the funding supported by National Key $\text{R\&D}$ Program of China (No. 2018AAA0102004), National Natural Science Foundation of China (No. 62141607, U1936219).

\nocite{langley00}

\bibliography{example_paper}
\bibliographystyle{icml2022}

\newpage
\appendix
\onecolumn
\icmltitle{Supplementary Materials: \\Model Agnostic Sample Reweighting for Out-of-Distribution Learning}
This appendix can be divided into the following parts:
\begin{enumerate}
    \item Section \ref{dataset} gives the details of datasets in IRM and DRO.
    \item Section \ref{exp} presents experimental configurations of this work.
    \item Section \ref{sec:hist} presents experiments on weight distributions of different groups to show the ability of MAPLE to identify weights for each training samples.
    \item Section \ref{sec:sparsity} presents experiments on the effectivenss of improving training speed via sparsity constraint on training sample size.
    \item Section \ref{trans} presents experiments on validation of transferability of sample weights.
    \item Section \ref{app:proof:thm:Identifiability} presents proof of Theorem \ref{thm:Identifiability}
    \item Section \ref{app:proof:thm:finite_bound} presents proof of Theorem \ref{thm:finite_bound}
    \item Section \ref{sec:bilevel_related} introduces related works on bi-level optimization.
    \item Section \ref{sec:future_works} presents discussions on future works.
\end{enumerate}

\section{Dataset Details} \label{dataset}
ColoredMNIST is the most widely used benchmark in IRM and ColoredObject, CIFARMNIST are adopted to showcase the superior performance of MAPLE on more challenging largescale settings. The labels for IRM datasets are generated from the core features. The spurious features have strong correlations with the labels in the training set but the correlation reverses in the testing set. In each dataset there exist two training environments and one testing environment with different correlations. We combine the correlations of two training environments and one testing environment into a correlation tuple. Label noise is added to the datasets to make the task more challenging \cite{Arjovsky2019Invariant, zhang2021can}. 

Waterbirds and CelebA are real-world datasets and we adopt them to demonstrate the generalizability of MAPLE to real-world scenarios. Waterbirds and CelebA are both binary prediction tasks. In each dataset, there exists a binary spurious feature highly correlated with the label. We follow the challenging setting of \citet{liu2021just} that no group annotation is provided in the training dataset and group annotations are provided in the small validation set. 

\textbf{ColoredMNIST} \cite{Arjovsky2019Invariant}.\label{data:cmnist} It contains images from MNIST and the images are labeled as 0 or 1. Each image is attached with a color as the spurious feature. Correlation tuple is $(0.9, 0.8, 0.1)$. Noise ratio is 25\%. 

\textbf{ColoredObject} \cite{ahmed2020systematic, zhang2021can}. It is constructed by extracting 8 classes of objects from MSCOCO and put them onto colored backgrounds. Correlation tuple is $(0.999, 0.7, 0.1)$. Noise ratio is 5\%. 

\textbf{CIFARMNIST} \cite{shah2020pitfalls, yong2021empirical}. It is constructed by concatenating images of CIFAR10 with MNIST. The CIFAR images are the invariant features and the MNIST images are the spurious features. Correlation tuple is $(0.999, 0.7, 0.1)$.  Noise ratio is 10\%. 

\textbf{Waterbirds} \cite{WahCUB_200_2011, sagawa2019distributionally}. The Waterbirds dataset contains two group of birds, i.e., $\{$waterbird, landbird$\}$. There are two kinds of background, i.e., $\{$water background, land background$\}$. The background type is spuriously correlated with the bird type.  No background annotation is provided in the training dataset. 

\textbf{CelebA} \cite{liu2015deep, sagawa2019distributionally}. In the CelebA dataset, the task is to predict hair color, $\{$blond, dark$\}$, based on the image input. The attribute gender, $\{$male, female$\}$, is spuriously correlated with the hair color. 

\begin{table}[h]
\caption{Illustration of each dataset. Core and Spurious stand for the core and spurious features, respectively. Spurious features are highly correlated with the label. However, the correlations are reversed in the testing samples to simulate the distributional shift.}
\vskip 0.15in
  \centering
  \resizebox{0.6\linewidth}{!}{
  \begin{tabular}{ c  c  c  c c}
    \toprule
    Dataset & Core & Spurious & Training &  Testing \\
    \midrule
    \makecell{ColoredMNIST}& Digit & Color & 
    \begin{minipage}[c]{0.15\columnwidth}
		\centering
		\raisebox{-.15\height}{\includegraphics[width=\linewidth]{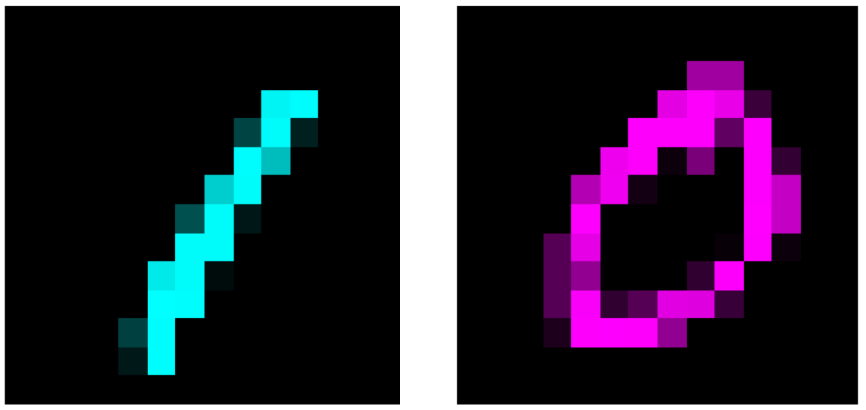}}
	\end{minipage}
& 
\begin{minipage}[c]{0.15\columnwidth}
		\centering
		\raisebox{-.15\height}{\includegraphics[width=\linewidth]{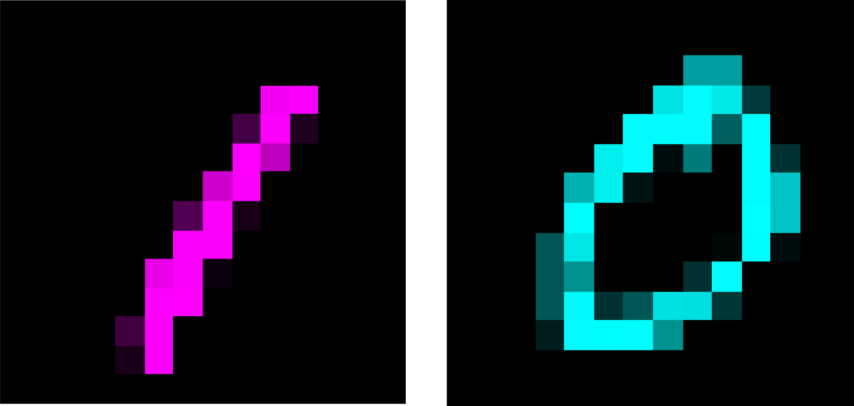}}
	\end{minipage} 
    \\ 
    \midrule
    ColoredObject & Object & Background & 
    \begin{minipage}[c]{0.15\columnwidth}
		\centering
		\raisebox{-.15\height}{\includegraphics[width=\linewidth]{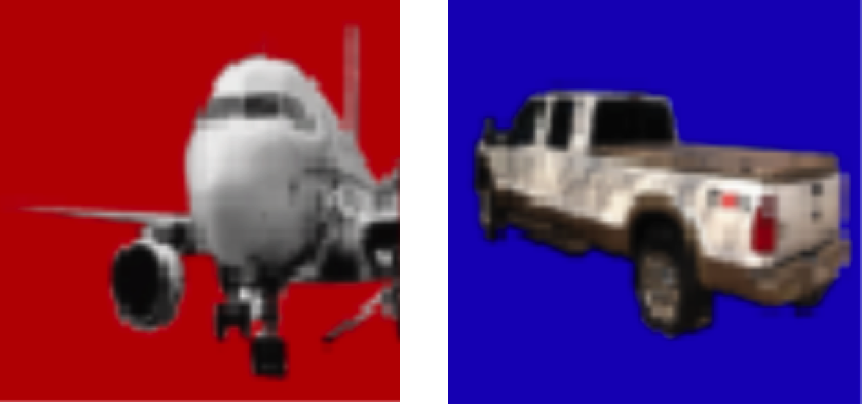}}
	\end{minipage}
&
\begin{minipage}[c]{0.15\columnwidth}
		\centering
		\raisebox{-.15\height}{\includegraphics[width=\linewidth]{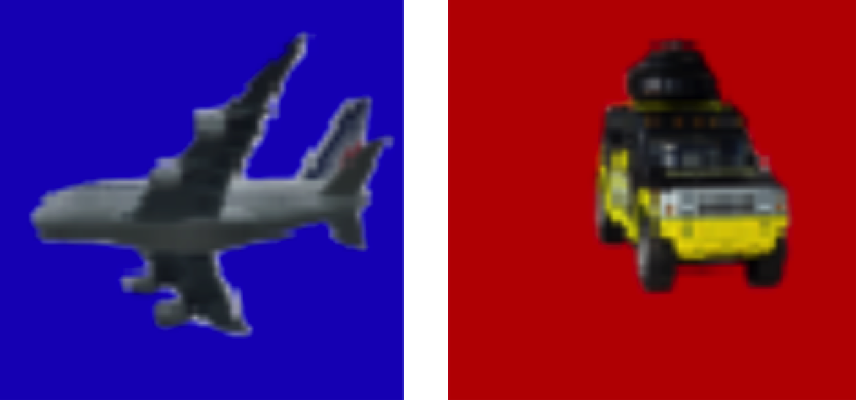}}
	\end{minipage}
    \\ 
    \midrule
CIFARMNIST& CIFAR & MNIST &  \begin{minipage}[c]{0.15\columnwidth}
		\centering
		\raisebox{-.15\height}{\includegraphics[width=\linewidth]{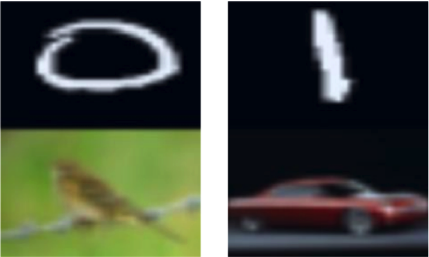}}
	\end{minipage} & \begin{minipage}[c]{0.15\columnwidth}
		\centering
		\raisebox{-.15\height}{\includegraphics[width=\linewidth]{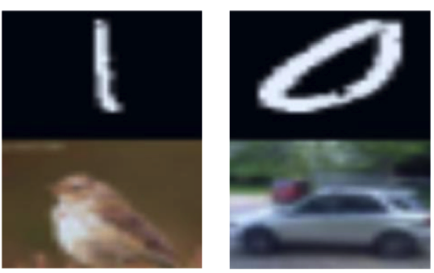}}
	\end{minipage}
	\\
	\midrule

Waterbirds& Bird & Background &  \begin{minipage}[c]{0.15\columnwidth}
		\centering
		\raisebox{-.15\height}{\includegraphics[width=\linewidth]{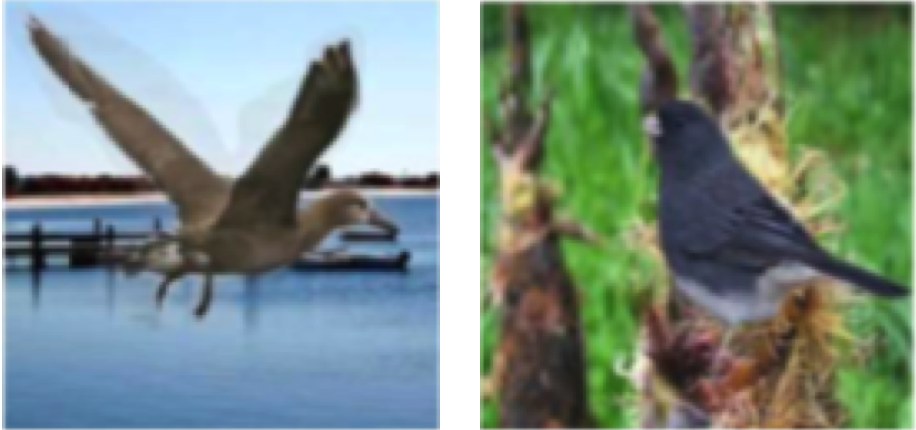}}
	\end{minipage} & \begin{minipage}[c]{0.15\columnwidth}
		\centering
		\raisebox{-.15\height}{\includegraphics[width=\linewidth]{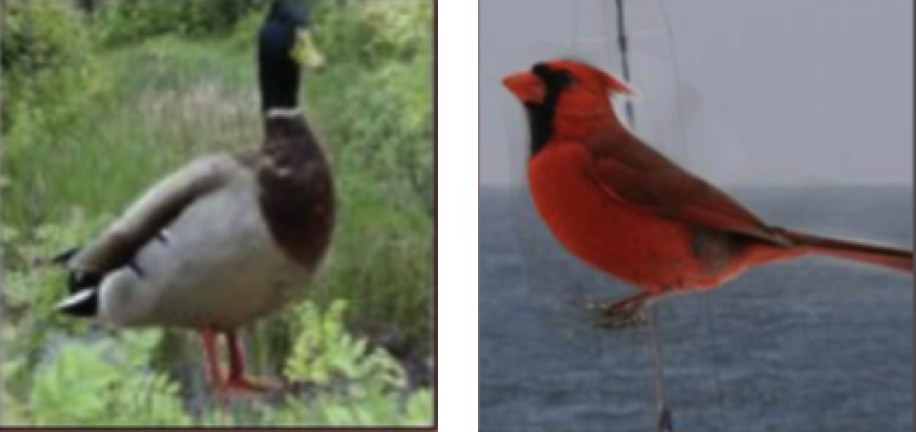}}
	\end{minipage}
	\\	
	\midrule

CelebA& Hair Color & Gender&  \begin{minipage}[c]{0.15\columnwidth}
		\centering
		\raisebox{-.15\height}{\includegraphics[width=\linewidth]{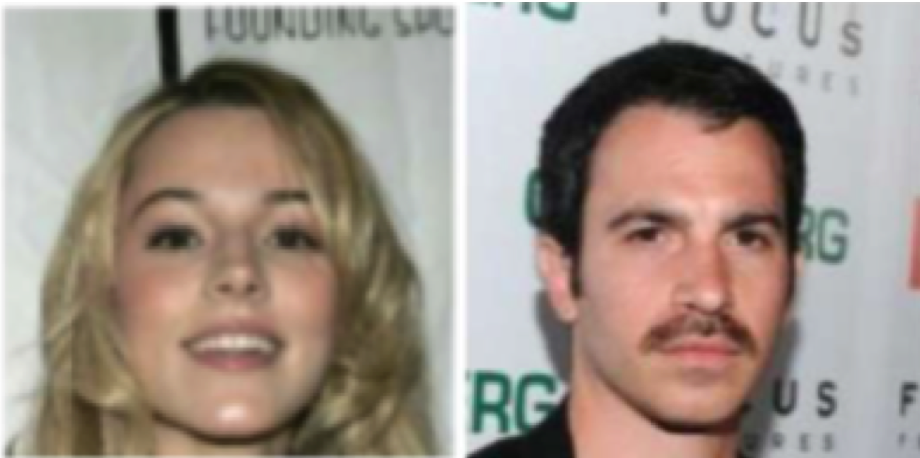}}
	\end{minipage} & \begin{minipage}[c]{0.15\columnwidth}
		\centering
		\raisebox{-.15\height}{\includegraphics[width=\linewidth]{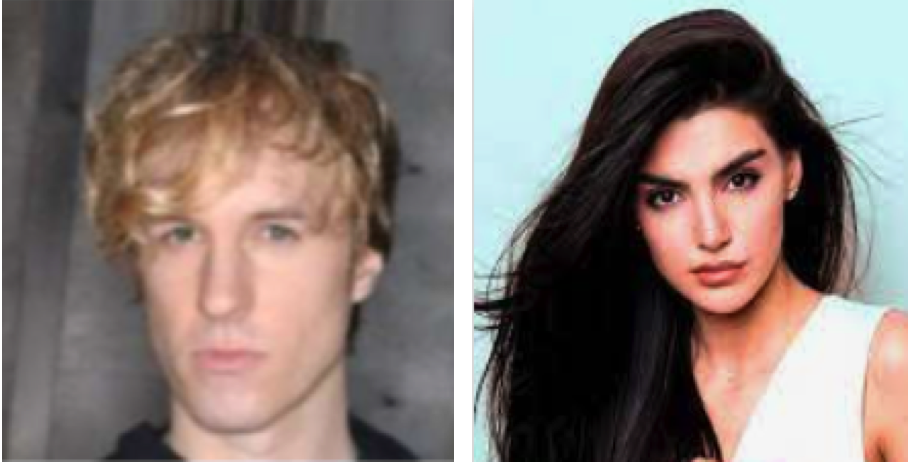}}
	\end{minipage}
	\\
    \bottomrule
  \end{tabular}
  }
  \vskip -0.1in
\end{table}
\newpage
\section{Experimental Configurations} \label{exp}

\begin{table}[H]
\caption{Experimental Configurations of MAPLE. The hyperparameters of sample weight and probability optimization are obtained via grid search on validation set on ColoredMNIST and applied directly to other scenarios. The demonstrates the robustness of MAPLE to different settings. We directly takes the regular training recipe for ERM training as the hyperparameters of inner loop model parameter optimization. We early stop in the inner loop as we find that training for such schedule is enough to obtain approximately best performance in validation set. 
} 
\begin{center}
\footnotesize
\begin{tabular}{p{3.0cm}<{\centering}p{2cm}<{\centering}p{2cm}<{\centering}p{2cm}<{\centering}p{2cm}<{\centering}p{2cm}<{\centering}}
\toprule
Dataset & ColoredMNIST & CIFARMNIST & ColoredObject & Waterbirds & CelebA\\ \cmidrule(){1-6}
GPUs & 1 & 1 & 1 & 1 & 8 \\ \cmidrule(){1-6}
Batch Size & 50000 & 1000 & 1000 & 128 & 1024 \\ \cmidrule(){1-6} 
Outer Iterations & 100 & 100 & 100 & 50 & 100 \\ \cmidrule(){1-6}
Inner Training Schedule& 100 iterations & 100 iterations & 100 iterations & 3 epochs & 1 epoch \\ \cmidrule(){1-6} 
Sample Weight Optimizer & Adam & Adam & Adam & Adam & Adam  \\ \cmidrule(){1-6}
Sample Weight Learning Rate & 0.25 & 0.25 & 0.25 & 0.25 & 0.25\\ \cmidrule(){1-6}
Sample Probability Optimizer & Adam & Adam & Adam & Adam & Adam \\ \cmidrule(){1-6}
Sample Probability Learning Rate & 5e-2 & 5e-2 & 5e-2 & 5e-2 & 5e-2 \\ \cmidrule(){1-6}
Model Parameter Optimizer & SGD & SGD & SGD & SGD & SGD\\  \cmidrule(){1-6}
Model Parameter Learning Rate & 1e-1 & 1e-2 & 1e-2 & 1e-4 & 1e-4\\  \cmidrule(){1-6}
Model Parameter Weight Decay & 1e-1 & 1e-2 & 1e-2 & 1e-1 & 1e-2 \\ \bottomrule
\end{tabular}
\end{center}
\end{table}

\section{Weight Distributions of Different Groups}\label{sec:hist}
\begin{figure}[t!] 
\begin{center}
\centering  
\includegraphics[scale=0.4]{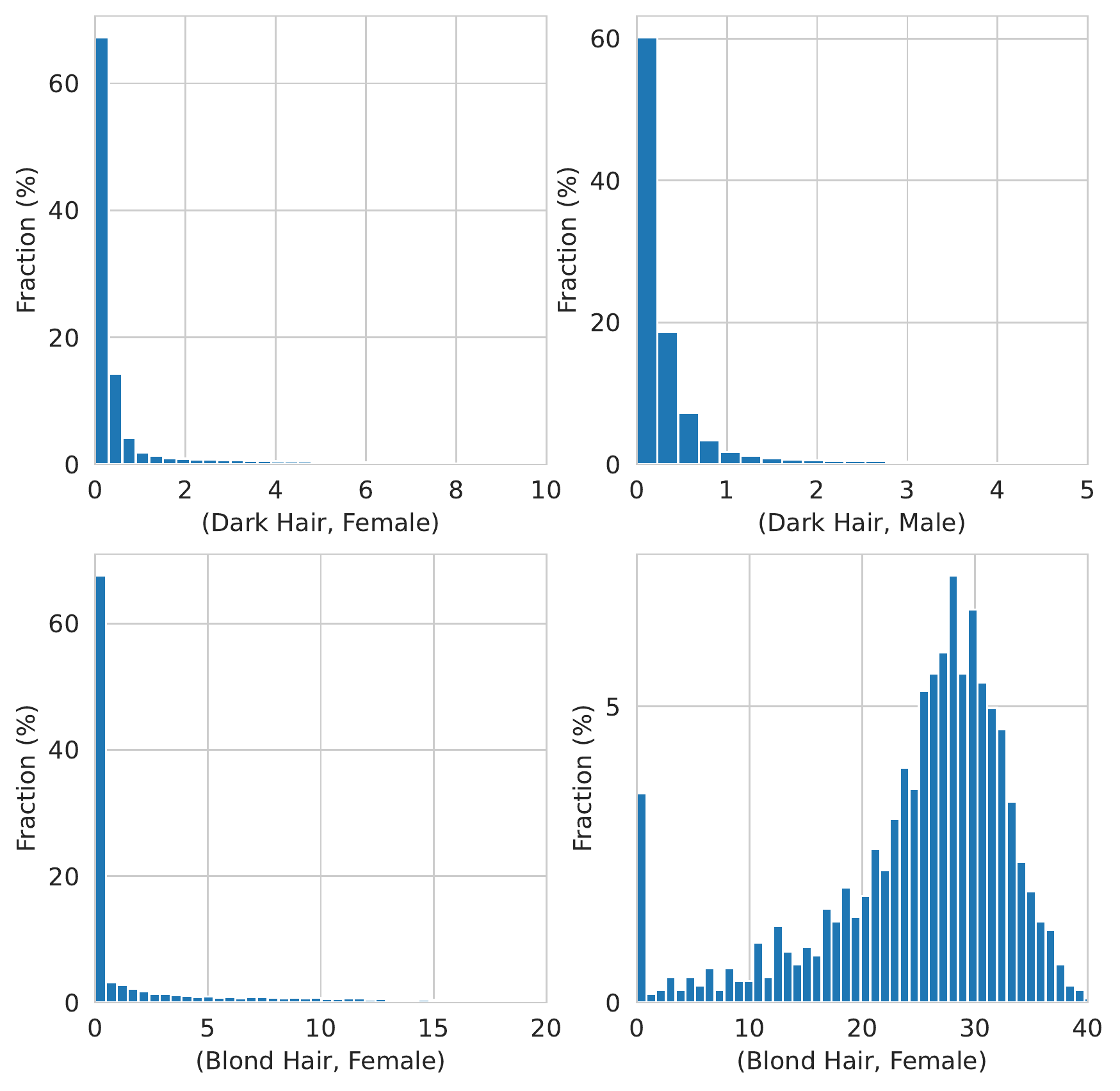}
\end{center}
\vspace*{-10pt}
\caption{Histogram of weights for four groups in CelebA. MAPLE automatically upweights the weights of group (Blond Hair, Female) and the histograms demonstrate that the weight assigned to different groups need not be the same.}
\vspace*{-5pt}
\label{fig:hist}
\end{figure}

We further plot the histogram of samples weights in Figure \ref{fig:hist} for four groups in CelebA experiment at the end of training. It indicates that the weights of group (Blond Hair, Female) flattens to around 30, while the weights of other groups still remainly lies around 1. This is consistent with our primal goal to upweight the group with few training samples, and MAPLE sucessfully achieve this without any training group annotations. We also discovers that the sample weight assigned to different training samples need not be the same. This demonstrates another reason why MAPLE beats JTT and Upweighting by a large margin.

\section{Effectiveness of Sparsity in Promoting Training Speed} \label{sec:sparsity}
Table \ref{tab:sparsity} demonstrates the comparison of training speed between MAPLE with no sparsity constraint on sample sizes and MAPLE. MAPLE saves a lot of inner loop computation time.

\begin{table}[htb!]
\caption{Comparing computational time of inner loop of different methods on Waterbirds. MAPLE(NS) indicates MAPLE with no sparsity constraint.}


\begin{center}
{\footnotesize
\begin{tabular}{p{2.3cm}<{\centering} p{1.4cm}<{\centering}p{1cm}<{\centering}}
\toprule
Method& MAPLE(NS) & MAPLE\\ \cmidrule(){1-3}
GPU Hours & 8.43 & \textbf{6.74}\\ \bottomrule
\end{tabular}
}
\end{center}
\label{tab:sparsity}
\end{table}

\section{Validation of Transferability of Sample Weights}\label{trans}
We transfer the sample weights searched via ResNet-18 and directly apply it to train the weighted training samples on ResNet-50. Table \ref{tab:agnostic} demonstrates that the searched sample weights on ResNet-18 can be successfully applied to perform weighted ERM training on ResNet-50, even with slight performance boost.

\begin{table}[htb!]

\caption{Validating transferability of sample weights on Waterbirds, from ResNet-18 on seaching phase and ResNet-50 on downstream weighted training phase.}
\begin{center}
{\footnotesize
\begin{tabular}{p{4cm}<{\centering} p{3cm}<{\centering}p{3cm}<{\centering}}
\toprule
Sample Weights Searched on ResNet-18& Weighted ERM Training on ResNet-18 & Weighted ERM Training on ResNet-50\\ \cmidrule(){1-3}
Worst-group Acc&  91.2\% & \textbf{91.6}\% \\ \bottomrule
\end{tabular}
}
\end{center}
\label{tab:agnostic}
\end{table}

\section{Proof of Theorem \ref{thm:Identifiability}}
\label{app:proof:thm:Identifiability}
By Assumption \ref{ass:strict_pos}, $\bbP(\by,\bz_c, \bz_s)>0$. Then we can define the following weighting function 
\begin{align}
    \label{eqn:desired_wegight}
    w(\by,\bx) :=  \frac{\bbP(\by,\bz_c)\bbP(\bz_s)}{\bbP(\by,\bx)}
\end{align}
Below, we will show that $w(\bx,y)$ is the desired weight function, and the solution of this ordinary least square re-weighted by $w(\bx,y)$ is the optimal debiased predictor $\bs{\bar \theta}$. Specifically, 
\begin{align}
    \label{eqn:weighted_loss}
     \cL(\bs \theta; w)  &= \int (y - \bx^\intercal \bs \theta)^2 \bbP_w(\bx, y) d \bx dy,
 \end{align}
 It is easy to know that the minimizer of Eqn. \eqref{eqn:weighted_loss}.
 \begin{align*}
     \bs{\theta}^*(w) & = \left(\int \bx \bx^\top \bbP_w(\bx, y) d \bx dy\right)^{-1} \int \bx y \bbP_w(\bx, y) d \bx dy \\
     & = \left(\int \bs S \bz \bz^\top \bs{S}^\top \bbP_w(\bx, y) d \bx dy\right)^{-1} \int \bs{S} \bz y \bbP_w(\bz, y) d \bx dy \\
     & = (\bs{S}^\top)^{-1} \left(\int  \bz \bz^\top  \bbP_w(\bz, y) d \bz dy\right)^{-1} \int  \bz y \bbP_w(\bz, y) d \bz dy \\
     & = (\bs{T}^\top) \left(\int  \bz \bz^\top  \bbP_w(\bz, y) d \bz dy\right)^{-1} \int  \bz y \bbP_w(\bz, y) d \bz dy.
 \end{align*}
At last, we are going to show $\left(\int  \bz \bz^\top  \bbP_w(\bz, y) d \bz dy\right)^{-1} \int  \bz y \bbP_w(\bz, y) d \bz dy$ will be equal to $[\bs{\bar \theta}_c; \mathbf{0}]$ as defined in Definition \ref{defi:optimal_debiased_predictor}.
\begin{proof}

We denote $\Sigma^w = \int \bx \bx^\top \bbP_w(\bx, y) d \bx dy$, and turn to simplify $\theta^{*}(w)$ by computing $\Sigma^{w}$ and $Cov^w$

It follows that 
\begin{align*}
      \bbP_w(\by, \bz_c, \bz_s) =\bbP_w(\by, \bx)=  w(\by, \bx) \bbP(\by,\bx) = \bbP(\by,\bz_c)\bbP(\bz_s).
\end{align*}
It is easy to see $\bbP_w(\by, \bz_c) = \bbP(\by,\bz_c)$ and $\bbP_w(\bz_s) = \bbP(\bz_s)$ because  
\begin{align*}
    &\bbP_w(\by, \bz_c) = 
    \int_{\bz_s} \bbP_w(\by, \bz_c, \bz_s)  =  \int_{\bz_s} \bbP(\by,\bz_c)\bbP(\bz_s) =  \bbP(\by,\bz_c)\int_{\bz_s} \bbP(\bz_s)  = \bbP(\by,\bz_c)\\
    &\bbP_w(\bz_s)  = 
    \int_{\by, \bz_c} \bbP_w(\by, \bz_c, \bz_s) =  \int_{\by, \bz_c} \bbP(\by,\bz_c)\bbP(\bz_s)=  \bbP(\bz_s) \int_{\by, \bz_c}  \bbP(\by,\bz_c) = \bbP(\bz_s)
\end{align*}

So we further have 
$$P_{w}(\by,\bz_c, \bz_s)=P(\by,\bz_c)P(\bz_s) = \bbP_w(\by, \bz_c) \bbP_w(\bz_s).$$ It also leads to $$P_{w}(\bz_c, \bz_s) = \bbP_w(\bz_c) \bbP_w(\bz_s)$$  
$$P_{w}(\by, \bz_s) = \bbP_w(\by) \bbP_w(\bz_s)$$

It follows that  
\begin{align*}
    \Sigma_c^w & := \bbE[w(\bx, y) \bz_c  \bz_c^\intercal ] = \int \bz_c  \bz_c^\intercal \bbP_w(\bx, y) = \int \bz_c  \bz_c^\intercal \bbP_w(\bz, y) =  \int \bz_c  \bz_c^\intercal \bbP_w(\bz_c) = \int \bz_c  \bz_c^\intercal \bbP(\bz_c) = \Sigma_c \\
    \Sigma_b^w & := \bbE[w(\bx, y) \bz_s  \bz_s^\intercal ] = \int \bz_s  \bz_s^\intercal \bbP_w(\bx, y) = \int \bz_s  \bz_s^\intercal \bbP_w(\bz, y) =  \int \bz_s  \bz_s^\intercal \bbP_w(\bz_s) = \int \bz_s  \bz_s^\intercal \bbP(\bz_s) = \Sigma_b
\end{align*}
Furthermore, 
\begin{align*}
    & \mbox{Cov}^w(\bz_c, \bz_s)\\
    & = \bbE[w(\bx, y)\bz_c^\intercal \bz_s] -\bbE[w(\bx, y)\bz_c]^\intercal \bbE[w(\bx, y)\bz_s] \\
    & = \int \bbP_w(\bz_c, \bz_s) \bz_c^\intercal \bz_s d\bz_c d\bz_s - \bbE[w(\bx, y)\bz_c]^\intercal \bbE[w(\bx, y)\bz_s] \\
    & = \int \bbP_w(\bz_c) \bbP_w(\bz_s) \bz_c^\intercal \bz_s d\bz_c d\bz_s  - \bbE[w(\bx, y)\bz_c]^\intercal \bbE[w(\bx, y)\bz_s] \\
    & =  \bbE[w(\bx, y)\bz_c]^\intercal \bbE[w(\bx, y)\bz_s]  - \bbE[w(\bx, y)\bz_c]^\intercal \bbE[w(\bx, y)\bz_s] = \mathbf{0}.
\end{align*}
Similarly, we can obtain 
\begin{align*}
    \bbE[w(\bx, y) \bz_c  y] & = \bbE[ \bz_c  y], \quad \bbE[w(\bx, y) \bz_s  y] = \mathbf{0}.
\end{align*}
Putting these together, we have 
\begin{align*}
    & \Sigma^w =  \begin{bmatrix}
    \Sigma_c^w & \mbox{Cov}^w(\bz_c, \bz_s) \\
    \mbox{Cov}^w(\bz_s, \bz_c) & \Sigma_s^w
    \end{bmatrix} = 
    \begin{bmatrix}
    \Sigma_b & \mathbf{0} \\
    \mathbf{0} & \Sigma_c
    \end{bmatrix}, \\
    & \bbE[w(\bx, y) \bz  y]  =  
    \begin{bmatrix}
    \bbE[w(\bx, y) \bz_c  y] \\
    \bbE[w(\bx, y) \bz_s  y]
    \end{bmatrix}  =  
    \begin{bmatrix}
    \bbE[ \bz_c  y] \\
    0
    \end{bmatrix}.
    \end{align*}

Then 
\begin{align*}
    \bs \theta^*(w) & = (\Sigma^w)^{-1} \bbE[w(\bx, y) \bz  y] = \begin{bmatrix}
    \Sigma_c & \mathbf{0} \\
    \mathbf{0} & \Sigma_b
    \end{bmatrix}^{-1} 
    \begin{bmatrix}
    \bbE[ \bz_c  y] \\
    0
    \end{bmatrix}  =  
    \begin{bmatrix} 
    \Sigma_c^{-1}\bbE[ \bz_c  y]\\
    0
    \end{bmatrix} = \begin{bmatrix} 
    \bs{\bar \theta}_c\\
    0 
    \end{bmatrix}= \bs{\bar \theta}.
\end{align*}
The second part proof is straightforward. By Assumption \ref{ass:identifibility}, for any $\theta \neq \theta^*=\theta_w$, we have
\begin{align}
    \label{aux_eqn:unqiueness_of_theta_w}
    \cR(\theta) > \cR(\bs {\bar \theta}) =  \cR(\bs{\theta}^*(w)).
\end{align}
We already know that $\bs{\theta}^*(w)$ is in the feasible solution of MAR. Eq. \eqref{aux_eqn:unqiueness_of_theta_w} further shows that $\bs{\theta}^*(w)$ achieves the minimum loss of $\cR$. Putting these together, we conclude that MAR uniquely identify $\bar \theta$.
\end{proof}
\section{Proof of Theorem \ref{thm:finite_bound}}
\label{app:proof:thm:finite_bound}
By the bounded difference inequality (Corollary 2.21 of \cite{wainwright2019high}), given any $\bw$, we have with probability $1-\delta/2$,
\begin{align}
    \label{eqn:basic_inequality}
    \small
    \cR(\bs{\hat \theta}^*(\bw);  \cD_v) \leq  \cR(\bs{\hat \theta}^*(\bw)) + M\sqrt{\frac{\ln(2/\delta)}{2N}},
\end{align}
where $\cR(\bs{\hat \theta}^*(\bw);  \cD_v)$ is the OOD risk on the validation dataset $\cD_v$ and $\cR(\bs{\hat \theta}^*(\bw))$ is the population OOD risk. Then we have with probability $1 - \delta$,

\begin{align*}
& \cR(\bs{\hat \theta}^*(\hat \bw)) \\
\leq & \cR(\bs{\hat \theta}^*(\hat \bw); \cD_v) + M\sqrt{\frac{2\ln(2|\cW|/\delta)}{N}} \\
    \leq &   \cR(\bs{\hat \theta}^*(\bw);  \cD_v)+ M\sqrt{\frac{\ln(2|\cW|/\delta)}{2N}} + \epsilon \\
    \leq & \cR(\bs{\hat \theta}^*(\bw)) + M\sqrt{\frac{\ln(2/\delta)}{2N}} + M\sqrt{\frac{\ln(2|\cW|/\delta)}{2N}} + \epsilon \\
    \leq & \cR(\bs{\hat \theta}^*(\bw)) + M\sqrt{\frac{2\ln(2|\cW|/\delta)}{N}} + \epsilon ,
\end{align*}

The first inequality because we require inequality \eqref{eqn:basic_inequality} to hold uniformly for all $|\cW|$ functions. The second inequality is because $\hat \bw$ is the $\epsilon$-approximated solution descrided in Eqn. \eqref{eqn:epsilon}. The third inequality is applying inequality \eqref{eqn:basic_inequality}. The forth inequality is because $|\cW|>1$. Taking infimum over $\bw$ on the right hand side, we obtain the desired bound.

\section{Related Works on Bilevel Optimization} \label{sec:bilevel_related}
Bilevel optimization \cite{sinha2017review} has aroused much attention in recently due to its ability to handle hierarchical decision making processes. Previous works utilize bilevel optimization in multiple areas of research, such as hyper-paramter optimization \cite{lorraine2020optimizing, maclaurin2015gradient,pedregosa2016hyperparameter, mackay2019self}, meta learning \cite{finn2017model, nichol2018reptile}, neural architecture search \cite{liu2018darts, xu2019pc, shi2020bridging, yao2021joint, gao2021autobert, yao2021g,shi2021sparsebert} and sample re-weighting \cite{ren2018learning, shu2019meta}, coreset selection \cite{coreset, borsos2020coresets}.

\section{Future Directions} \label{sec:future_works}
MAPLE stills needs to demonstrate its applicability to NLP tasks especially on today's large pretraining language models \cite{devlin2018bert, radford2019language, liu2019roberta, diao2019zen, brown2020language}, cross-modal tasks \cite{gu2018look, gao2022unison, https://doi.org/10.48550/arxiv.2205.15237}, domain adaptation tasks \cite{diao2021taming, huang2022arch} and self-supervised learning tasks \cite{he2020momentum,grill2020bootstrap,chen2021multisiam,liu2022task}. It is also interesting to explore how MAPLE interacts with other parallel domain generalization methods \cite{luo2018towards, bai2021decaug, bai2021ood} , how it interacts with other methods focusing on model sparsity \cite{Shao_2019_CVPR, sparseirm, shi2021sparsebert} and how it performs on more challenging benchmarks \cite{ye2022ood}.

\end{document}